\documentclass[11pt]{article}

\usepackage[preprint]{acl}

\usepackage{times}
\usepackage{latexsym}

\usepackage[T1]{fontenc}

\usepackage[utf8]{inputenc}

\usepackage{microtype}

\usepackage{inconsolata}

\usepackage{graphicx}
\usepackage{bm}
\usepackage{amsmath}
\usepackage{amsfonts}
\usepackage{amssymb}
\usepackage{algorithm}
\usepackage{algpseudocode}
\usepackage{amsmath} 
\usepackage{mathtools}
\usepackage{hyperref}
\usepackage{booktabs}
\usepackage{amssymb}
\usepackage{microtype}
\usepackage{url}
\usepackage{enumitem}
\usepackage{makecell}
\usepackage{url}
\usepackage[labelformat=simple]{subcaption}
\usepackage{CJKutf8}
\usepackage{latexsym}
\usepackage{amsfonts}
\usepackage{amsmath}
\usepackage{multicol}
\usepackage{multirow}
\usepackage{xspace}
\usepackage{booktabs}
\usepackage{array}
\usepackage{threeparttable}
\usepackage{tabularx}
\usepackage{enumitem}
\usepackage{xcolor,colortbl}
\usepackage{color}
\usepackage{setspace}
\usepackage{makecell}
\usepackage{listings}
\usepackage{xltabular}
\usepackage{tcolorbox}
\tcbuselibrary{skins, breakable, listings}
\usepackage{arydshln}
\usepackage{caption}
\usepackage{marvosym}
\usepackage{fontawesome}
\usepackage{cleveref}

\definecolor{deepred}{RGB}{160, 0, 0}
\definecolor{acl-blue}{RGB}{115, 170, 220}     
\definecolor{acl-green}{RGB}{46, 139, 87}
\definecolor{fail-red}{RGB}{200, 60, 60}          
\definecolor{fail-bg}{RGB}{255, 245, 245}         
\definecolor{fail-dark}{RGB}{139, 0, 0}           
\definecolor{insight-bg}{RGB}{255, 250, 220}
\definecolor{insight-border}{RGB}{210, 160, 30}
\definecolor{bg-thought}{RGB}{250, 251, 252}   
\definecolor{bg-code}{RGB}{250, 250, 250}      
\definecolor{border-gray}{RGB}{220, 220, 220}  

\lstdefinestyle{normalcode}{
    language=Python,
    basicstyle=\ttfamily\scriptsize\color{black!80}, 
    keywordstyle=\color{blue!60!black}\bfseries,
    stringstyle=\color{orange!60!black},
    commentstyle=\color{green!40!black}\itshape,
    numberstyle=\tiny\color{gray},
    numbers=left,
    stepnumber=1,
    frame=none,
    backgroundcolor=\color{bg-code},
    breaklines=true,
    tabsize=4,
    showstringspaces=false,
    aboveskip=2pt, 
    belowskip=2pt
}

\newtcolorbox{thoughtblock}[1][]{
    enhanced, frame hidden, colback=bg-thought,
    borderline west={3pt}{0pt}{acl-blue!60},
    title={\small $\diamondsuit$ \ \textbf{Reasoning}},
    coltitle=acl-blue!90!black, fonttitle=\bfseries,
    attach boxed title to top left={xshift=0mm, yshift=0mm},
    boxed title style={boxrule=0pt, colframe=white, colback=white},
    left=4mm, right=2mm, top=0mm, bottom=2mm,
    fontupper=\small,
    #1
}

\newtcolorbox{codeblock}[1][]{
    enhanced, colback=bg-code, colframe=border-gray, boxrule=0.5pt, arc=2pt,
    title={\small $\rhd$ \ \textbf{Action: Code Interpreter}},
    coltitle=black!70, colbacktitle=gray!10, fonttitle=\bfseries,
    attach boxed title to top left={xshift=2mm, yshift=-2mm},
    boxed title style={boxrule=0pt, colframe=white},
    top=3mm, bottom=1mm, left=1mm, right=1mm, #1
}

\newtcolorbox{obsblock}[1][]{
    enhanced, colback=black!85, colframe=black!85, arc=2pt,
    title={\small $\star$ \ \textbf{Observation}},
    coltitle=white!90, fonttitle=\bfseries,
    attach boxed title to top left={xshift=2mm, yshift=-3mm},
    boxed title style={boxrule=0pt, colframe=black!85},
    coltext=white, 
    fontupper=\ttfamily\scriptsize, 
    top=2mm, bottom=2mm, left=2mm, right=2mm, #1
}
\title{Learning from the Irrecoverable: Error-Localized Policy Optimization for Tool-Integrated LLM Reasoning}

\author{%
  Qiao~Liang$^{2}$\thanks{Work done during an internship at MYbank.}, 
  Yuke~Zhu$^{1}$,
  Chao~Ge$^{1}$,
  Lei~Yang$^{1}$,
  Ying~Shen$^{2}$,
  Bo~Zheng$^{1}$,
  Sheng~Guo$^{1}$\thanks{Corresponding author.}\\
  $^{1}$MYbank, Ant Group, China \\
  $^{2}$Tongji University, China \\
  \texttt{\{2333091, yingshen\}@tongji.edu.cn} \\
  \texttt{\{felix.yk, guosheng.guosheng\}@mybank.cn}
}

\begin{document}
\maketitle
\begin{abstract}
Tool-integrated reasoning (TIR) enables LLM agents to solve tasks through planning, tool use, and iterative revision, but outcome-only reinforcement learning in this setting suffers from sparse, delayed rewards and weak step-level credit assignment.
In long-horizon TIR trajectories, an early irrecoverable mistake can determine success or failure, making it crucial to localize the \emph{first irrecoverable step} and leverage it for fine-grained credit assignment.
We propose \textbf{Error-Localized Policy Optimization (ELPO)}, which localizes the first irrecoverable step via binary-search rollout trees under a fixed rollout budget, converts the resulting tree into stable learning signals through hierarchical advantage attribution, and applies error-localized adaptive clipping to strengthen corrective updates on the critical step and its suffix.
Across TIR benchmarks in math, science QA, and code execution, ELPO consistently outperforms strong Agentic RL baselines under comparable sampling budgets, with additional gains in Pass@K and Major@K scaling, rollout ranking quality, and tool-call efficiency.
Our code will be publicly released soon.
\end{abstract}

\section{Introduction}


Reinforcement Learning with Verifiable Rewards (RLVR) substantially improves LLM performance on complex reasoning tasks by biasing generation toward correct solution trajectories~\cite{PPO,DPO,GRPO,wen2025reinforcement,zhang2025survey}.
Yet evidence suggests RLVR mainly reweights behaviors already present in the base model and may not reliably expand its reasoning boundary~\cite{yue2025does}.
This motivates agentic reinforcement learning (Agentic RL), where LLMs solve tasks by planning, invoking external tools, and iteratively revising intermediate decisions rather than relying solely on internal chain-of-thought generation~\cite{zhang2025landscape}.

\begin{figure}[t]
  \centering
  \includegraphics[width=1\columnwidth]{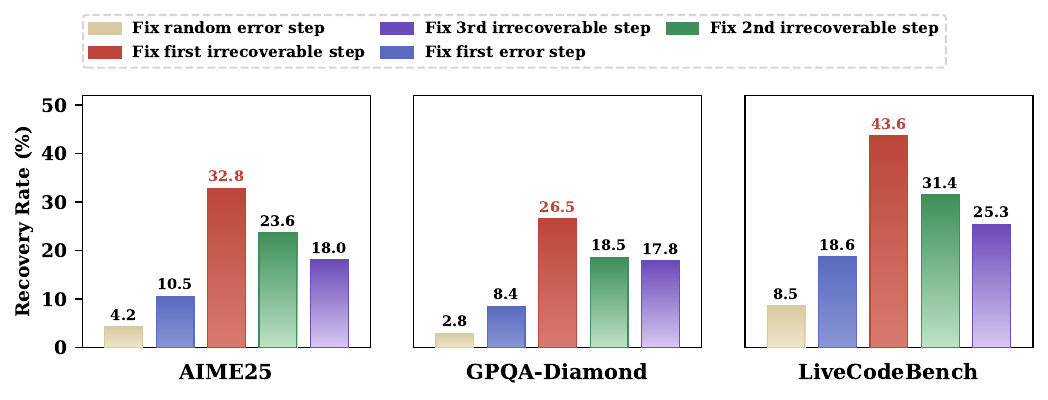}
  \caption{For each initially failed trajectory (Pass@16$=0$), we edit exactly one step (random error, first error, or the 1st/2nd/3rd irrecoverable step) and resample completions. Recovery is counted when the repaired run reaches Pass@16$=1$.}
  \vspace{-1.2em}
  \label{fig:intro}
\end{figure}

Tool-integrated reasoning (TIR), though enhances LLM's capability, brings new challenges to current outcome reward based reinforcement techniques, represented by GRPO~\cite{GRPO}. 
These outcome-based methods introduce sparsity and delay in reward signals.
The problems are less pronounced in general reasoning tasks but are significantly amplified in Agentic RL~\cite{landscapeofagenticrl}.
First, agentic systems rely on open-ended autonomous exploration, which dramatically enlarges the action and tool-use space.
Sparse terminal rewards therefore exacerbate the exploration bottleneck.
Second, agentic decision-making typically involves long, multi-step and hierarchical planning, where delayed outcome rewards fail to provide effective credit assignment across different stages and steps.
More critically, relying solely on outcome supervision can trap the agent in a repeated “trial–error–no-feedback” loop, making it difficult to ensure stable and directed learning. 
In summary, we argue that designing finer-grained process reward signals is crucial for Agentic RL.

Most of the recent works on Agentic RL spend their effort on outcome-level reward design~\cite{Search-R1, ToRL, ToolRL, OTC-PO, Artist, IKEA, AutoTIR, ASPO, AEPO, ARPO}, 
while only a few address the sparsity of process reward signals~\cite{GIGPO,Tree-GRPO,SPA-RL}. 
Overall, enabling fine-grained process supervision and reliable credit assignment in long-horizon TIR remains under-explored, which in turn constrains further progress in Agentic RL.

In this work, we directly confront the core of the problem and propose \textbf{Error-Localized Policy Optimization (ELPO)}, a novel approach that automatically identifies critical error steps during rollout and leverages them to provide finer-grained signals.
The motivation stems from a key experimental finding (shown in Figure~\ref{fig:intro}): localizing and correcting the \textbf{first irrecoverable error step} significantly increases the model's probability of producing the correct answer, whereas fixing a random error provides much less benefit. 
Thus, this motivates us to treat the recoverability of erroneous steps as a highly discriminative fine-grained signal.

However, incorporating this insight into Agentic RL training is non-trivial and requires addressing two key challenges. 
First, \textit{Budget bottleneck}: efficiently identifying irrecoverable error steps without exhaustive rollouts.
Second, \textit{Process signal generation}: deriving stable and discriminative supervision from localized errors.
To this end, we propose: (i) binary-tree-based Error Localization (BEL) (\cref{sec:bel}) with entropy-gap-guided path pruning for efficient error localization; (ii) a hybrid trajectory- and branch-level credit assignment scheme (\cref{sec:fga}) and (iii) error-localized adaptive clipping (\cref{sec:ecl}) to strengthen corrective updates on the localized critical step. 
With these techniques, our method consistently outperforms strong Agentic RL baselines across diverse TIR benchmarks.

In summary, the key contributions are as follows:
\begin{itemize}[itemsep=0pt,parsep=0pt,topsep=0pt,partopsep=0pt]
  \item We identify an \textbf{error-localization gap} in outcome-only long-horizon TIR: sparse terminal rewards fail to isolate the \textbf{first irrecoverable step}, yielding diffuse credit assignment and weak corrections.
  \item We propose \textbf{Error-Localized Policy Optimization (ELPO)}, which localizes the first irrecoverable step via a binary-search rollout procedure under a fixed rollout budget, and turns this signal into error-localized advantages and targeted policy updates.
  \item Across diverse TIR benchmarks, ELPO consistently outperforms strong Agentic RL baselines.
  Beyond final accuracy, analyses on Pass@k and Major@k, rollout ranking quality, and tool-call efficiency further validate its effectiveness and stability.
\end{itemize}

\section{Related Work}
\subsection{Tool-integrated Reasoning Agents}
Recent studies have explored enabling LLMs to act as decision-making agents equipped with external tools such as search engines, calculators, code interpreters, and APIs~\cite{ReAct, ToolFormer}.
Early approaches primarily relied on prompting strategies such as ReAct~\cite{ReAct} and Reflexion~\cite{Reflexion} to interleave reasoning with tool calls, while later work examined finetuning LLMs on tool-usage traces to improve robustness in multi-step reasoning scenarios~\cite{ToolLLM, ToolAlpaca}. 
This line of work highlights both the promise and the difficulty of training reliable tool-augmented agents, particularly when errors made early in the reasoning process propagate through subsequent steps.

\subsection{Reinforcement Learning for LLM Agents}
Reinforcement learning has emerged as a key technique for improving long-horizon reasoning in LLMs.
Group-based policy optimization methods such as GRPO~\cite{GRPO} and DAPO~\cite{DAPO} avoid explicit value networks by normalizing rewards among multiple rollouts for the same prompt, enabling large-scale RL training for reasoning, search, and tool-use tasks.
Building on this foundation, early Agentic RL systems such as ToRL~\cite{ToRL} and Search-R1~\cite{Search-R1} apply outcome-based RL to tool-augmented reasoning, showing that verifiable terminal supervision can elicit effective tool use (e.g., code execution and web search).
Subsequent work~\cite{IKEA, GIGPO, ASPO} improves learning efficiency and stability by introducing finer-grained reward design or advantage shaping on top of outcome supervision.

Tree-structured rollouts have also become an active direction for extracting richer training signals.
Tree-GRPO~\cite{Tree-GRPO} constructs a rollout tree via stochastic expansion and leverages the tree structure for fine-grained advantage estimation, but its random branching can be sensitive to the initial trajectory quality.
More recent methods~\cite{ARPO, AEPO} further use entropy-guided expansion to steer branching and encourage more diverse sampling under the same rollout budget.

Despite this progress, existing approaches largely remain outcome-driven and do not explicitly localize the \emph{first irrecoverable step} in long, interdependent tool-use trajectories.
As a result, even with step-aware heuristics or tree-based advantages, credit can still be spread across many non-causal steps, and the agent receives no direct signal about \emph{where} the trajectory first becomes irrecoverable under a fixed sampling budget, which limits how targeted and stable policy updates can be for long-horizon TIR.

\section{Preliminaries}
\subsection{Problem setup}
We consider a tool-augmented reasoning setting where an LLM acts as an agent to solve multi-step tasks given a task description $x \in \mathcal{X}$.
At each step $t = 1,\dots,T$, the agent observes a state $\bm{s}_t \in \mathcal{S}$ and produces a textual action $\bm{a}_t \in V^{n}$, where $V$ is the token vocabulary and $n$ is the maximum step length.
The environment executes the requested tool calls (e.g., code interpreter, search engine), returns the corresponding outputs, and deterministically updates the state to $\bm{s}_{t+1}$.
A full episode induces a trajectory $\tau = \{(\bm{s}_1, \bm{a}_1), \dots, (\bm{s}_T, \bm{a}_T)\}$, governed by an LLM policy $\pi_\theta(\bm{a}_t|\bm{s}_t, x)$.
Rewards are only provided at the end of the episode: a terminal scalar reward $R(\tau) \in \mathbb{R}$ is assigned based on the correctness of the final answer.

\begin{figure*}
\centering
  \includegraphics[width=1\textwidth]{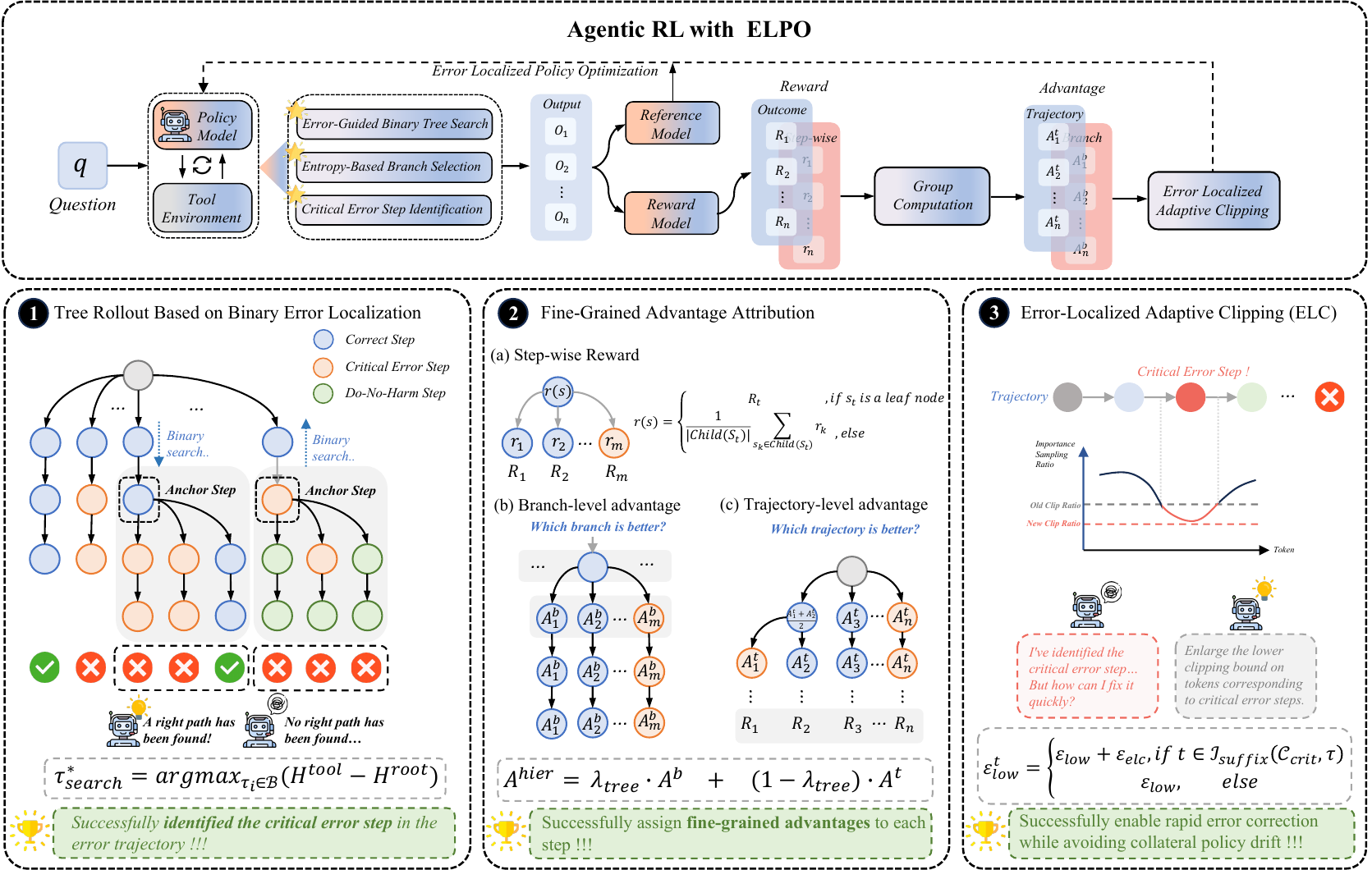}
  \caption{The overview of Error-Localized Policy Optimization (ELPO).}
  \label{fig:overallframework}
\end{figure*}

\subsection{Group-based Agentic RL}
Recent RL works converge on a group-based training recipe.
For each task input $x$ and tool set $\mathcal{T}$, the old policy
$\pi_{\theta_{\text{old}}}$ samples a group of trajectories
$\{o_i\}_{i=1}^G$, each with an outcome-only terminal reward $R_i$.
The advantage of trajectory $o_i$ is computed by normalizing rewards within
the group using
$
\hat{A}_i
= \frac{R_i - \mathrm{mean}(\{R_j\}_{j=1}^G)}
       {F_{\text{norm}}(\{R_j\}_{j=1}^G)}.
$

The GRPO-style objective is
{\small
\begin{equation}
\label{eq:grpo-obj}
\max_\theta \;
\mathbb{E}_{x,\{o_i\}}\bigg[
  \frac{1}{G}\sum_{i=1}^G
  \frac{1}{\sum_t m_{i,t}} \sum_t m_{i,t}\,
  L_{\text{CLIP}}(\theta,i,t)
\bigg],
\end{equation}
}
where $m_{i,t} \in \{0,1\}$ masks out environment feedback tokens so that
the loss only accumulates over agent-generated tokens.
The clipped surrogate loss follows PPO-style:
\begin{equation}
\label{eq:clip}
\small{
\begin{aligned}
L_{\text{CLIP}}(\theta,i,t)
  &= \min\Big(
      \rho_{i,t}(\theta)\hat{A}_i,\; \\[2pt]
  &\qquad
      \mathrm{clip}\big(
        \rho_{i,t}(\theta), 1-\varepsilon_{low}, 1+\varepsilon_{high}
      \big)\hat{A}_i
    \Big)
\end{aligned}
}
\end{equation}
with token-level importance ratio $\rho_{i,t}(\theta)
= \frac{\pi_\theta(o_{i,t}\mid o_{i,<t}, x; \mathcal{T})}
       {\pi_{\theta_{\text{old}}}(o_{i,t}\mid o_{i,<t}, x; \mathcal{T})}$.

\section{Methodology}
To address sparse terminal rewards in TIR, which assign the same outcome signal to all steps and leave the failure-causing decision unidentified, we propose Error-Localized Policy Optimization (ELPO).
ELPO builds a binary-search rollout tree to localize the first irrecoverable step with only $O(\log K)$ anchor probes, enabling targeted advantage attribution and updates.
As shown in Figure~\ref{fig:overallframework}, ELPO consists of three core designs:
\begin{enumerate}
  \renewcommand{\labelenumi}{(\theenumi)}
    \item \textbf{Tree Rollout Based on Binary Error Localization}: ELPO localizes the first irrecoverable step via tree rollouts with binary-search budget control.
  \item \textbf{Fine-Grained Advantage Attribution}: ELPO uses the rollout tree to assign per-step advantages from local branch comparisons and global trajectory ranking, reducing diffuse credit under outcome-only rewards.
  \item \textbf{Error-Localized Adaptive Clipping (ELC)}: Using the localized error step on the rollout tree, ELPO relaxes the clipping constraint for the error step and its generated suffix, allowing stronger corrective updates while preserving policy updating stability.
\end{enumerate}

\subsection{Tree Rollout Based on Binary Error Localization (BEL)} 
\label{sec:bel}
In ELPO, we use a binary-search rollout tree to pinpoint the first irrecoverable step in a failed trajectory.
Concretely, given a failed trajectory, we repeatedly select the middle step as an \textbf{anchor step}.
The anchor step and all preceding steps are fixed as a prefix, and several completions are sampled for the remaining suffix.
If any suffix completion succeeds, the prefix is recoverable and we search the later half; otherwise it is irrecoverable and we search the earlier half.
We repeat this process until the range collapses to a single step, which localizes the first irrecoverable step with only $O(\log K)$ anchor probes rather than exhaustively testing every step.
The main practical challenge is budget control: binary localization requires extra suffix completions, but the per-input rollout budget must remain fixed.
We address this with two complementary mechanisms.

\noindent\textbf{Trajectory Selection by Entropy Gap}.
Under a fixed budget, we run localization on only one informative failed trajectory per cycle because many failures are redundant and provide little additional contrast for learning.
We maintain a buffer $\mathcal{B}$ that stores the most recent erroneous trajectories from the current sampling.
Guided by recent study~\cite{ARPO}'s observation that token entropy typically spikes after tool-call feedback, we use the entropy gap between the tokens generated right after the task prompt and those generated right after tool-call feedback to decide which error trajectories deserve further branching.
For each trajectory $\tau \in \mathcal{B}$, we compute token entropy of the $i$ th token as
\begin{equation}
 H_i = - \sum_{j=1}^{V} p_{i,j} \log p_{i,j},\
\textbf{p}_i = \pi_{\theta}(\cdot \mid y_{<i}, x; \mathcal{T}),   
\end{equation}
and then compute $H^{\text{root}}$ as the average token entropy of the model’s initial response to the question, and $H^{\text{tool}}$ as the average entropy over all tool calls.
ELPO then selects a single trajectory
\begin{equation}
\tau_{\text{search}}^{*}
= \arg\max_{\tau_i \in \mathcal{B}} \big( H^{\text{tool}}(\tau_i) - H^{\text{root}}(\tau_i) \big),
\end{equation}
on which we subsequently run binary search. 
Once the selected error trajectory $\tau_{\text{search}}^{*}$ is used for branching, we clear the buffer and start accumulating new error samples for the next cycle.

\noindent\textbf{Adaptive Suffix Sampling at Anchor Steps}.
Anchor steps with larger post-tool uncertainty are more likely to be near the critical error.
We therefore allocate more suffix completions to anchors that are more likely to lie near the critical error, and probe low-uncertainty anchors with fewer samples.
Specifically, the number of suffix completions at the $m$-th anchor is set to
\begin{equation}
X_m = X_{\min} + (X_{\max}-X_{\min}) \cdot 
\sigma\!\big(\beta (H^{\text{tool}}_m - H^{\text{root}})\big),
\end{equation}
where $H^{\text{root}}$ is the entropy of the initial response and $H^{\text{tool}}_m$ the step-level entropy after the $m$-th tool call.
Large positive entropy gaps trigger more local sampling, while stable,
low-uncertainty steps are probed minimally.

Through this binary-search mechanism, ELPO identifies the first irrecoverable step in only $O(\log K)$ anchor evaluations, dramatically reducing rollout cost compared to naively checking all $K$ steps.
The pseudo-code is provided in Algorithm~\ref{alg:binary-error-localization} in Appendix~\ref{appendix:Pseudo Code}. 
An example of BEL rollout tree is provided in Appendix~\ref{example if BEL tree}.

\subsection{Fine-Grained Advantage Attribution}
\label{sec:fga}
BEL identifies \emph{where} a trajectory first becomes irrecoverable, but effective policy optimization further requires assigning credit \emph{how much} each decision contributes to success or failure.
Therefore, we leverage the binary search rollout tree to perform fine-grained advantage attribution.
We compute advantages from two sources: \textbf{branch-level} sibling comparisons and 
\textbf{trajectory-level} rollout ranking, and combine them into a hierarchical per-step advantage.
This two-granularity attribution yields localized yet globally consistent learning signals under outcome-only rewards.

\noindent\textbf{Branch-Level Advantage Attribution}.
Branch-level advantage is used to compare the relative quality of local branches under the same prefix.
To support branch comparisons, a \textbf{step-wise reward} is defined for each node in the rollout tree $\mathcal{G}_x=(\mathcal{S},\mathcal{E})$, estimating the expected terminal outcome of its prefix under the sampled continuations.
Let $R(s)$ denote the terminal reward when $s$ is a leaf node. The step-wise reward $r(s)$ is defined recursively:
\begin{equation}
\small{
r(s) =
\begin{cases}
R(s), & \text{if $s$ is a leaf node},\\[6pt]
\dfrac{1}{|\mathrm{Child}(s)|}\sum\limits_{c\in \mathrm{Child}(s)} r(c), & \text{otherwise},
\end{cases}
}
\label{eq:stepwise-reward}
\end{equation}
where $\mathrm{Child}(s)$ denotes the direct children of $s$ in $\mathcal{G}_x$.

For any branching node $s$ with $\mathrm{Child}(s)=\{c_1,\ldots,c_M\}$ and $M>1$, the branch-level advantage of each child branch $c_i$ is computed by normalizing $\{r(c)\}_{c\in \mathrm{Child}(s)}$:
\begin{equation}
A^{b}_{c_i}
= \frac{r(c_i) - \mu_b}{\sigma_b + \epsilon},
\end{equation}
where $\mu_b$ and $\sigma_b$ are the mean and standard deviation over $\{r(c)\}_{c\in \mathrm{Child}(s)}$.
This branch-level normalization provides localized credit signals exactly at decision points where alternative actions diverge.

\noindent\textbf{Trajectory-Level Advantage Attribution}.
\label{sec:traj-adv}
To provide a global preference signal by ranking complete rollouts for the same input, ELPO additionally defines a trajectory-level advantage over full trajectories.
For a given input $x$, the binary-search rollout tree $\mathcal{G}_x$ contains a set of $N$ completed leaf trajectories $\{\tau^{(i)}\}_{i=1}^{N}$, each associated with a terminal reward $\{R^{(i)}\}_{i=1}^{N}$.
Following group-based RL, we normalize these rewards within the same input and define
\begin{equation}
A^{t}(\tau^{(i)})=\frac{R^{(i)}-\mu_t}{\sigma_t+\epsilon},
\end{equation}
where $\mu_t$ and $\sigma_t$ are the mean and standard deviation of $\{R^{(i)}\}_{i=1}^{N}$.

In the rollout tree, an internal node $s$ may be shared by multiple trajectories.
We assign node-level trajectory credit by averaging the trajectory-level advantages passing through $s$:
\begin{equation}
A^{t}(s)=\frac{1}{|\mathcal{T}(s)|}\sum_{i\in\mathcal{T}(s)}A^{t}(\tau^{(i)}),
\end{equation}
where $\mathcal{T}(s)$ denotes the set of trajectories whose paths include $s$.
This term provides a coarse but stable signal on long linear segments, complementing the localized supervision from branch-level advantages.

\noindent\textbf{Hierarchical Advantage Attribution}.
\label{HAC}
Branch-level advantage localizes credit at branching decisions, while the trajectory-level term provides a stable global signal.
We combine them by a weighted sum:
\begin{equation}
A^{hier}(s)
=
\lambda_{\text{tree}}\, A^{b}(s)
+
(1-\lambda_{\text{tree}})\, A^{t}(s),
\end{equation}
where $\lambda_{\text{tree}}\in[0,1]$ controls the trade-off.
When $\lambda_{\text{tree}}=0$, ELPO reduces to trajectory-level group-based RL; when $\lambda_{\text{tree}}=1$, learning is driven purely by local branch comparisons.
We use $A^{\text{hier}}$ in the standard GRPO-style surrogate objective in place of the original group-level advantage.

\subsection{Error-Localized Adaptive Clipping (ELC)}
\label{sec:ecl}

Standard GRPO applies a uniform clipping ratio to all agent-generated tokens, which can limit the policy update strength even after BEL has identified first irrecoverable steps on the rollout tree.
Error-Localized Adaptive Clipping (ELC) relaxes the lower clipping bound for the critical step and its generated suffix.
This allows larger corrective updates on the error-related suffix while preserving GRPO’s clipping-based stability.

Concretely, we keep the PPO-style clipping form but replace the constant ratio lower bound $1-\varepsilon_{low}$ in Eq.~\eqref{eq:clip} with a token-dependent bound $1-\varepsilon_{\text{low}}^{t}$, where $\varepsilon_{\text{low}}^{t}$ is increased on the localized critical-step suffix:
\begin{equation}
\varepsilon_{\text{low}}^{t}=
\begin{cases}
\varepsilon_{low}+\varepsilon_{\text{elc}}, & t\in \mathcal{I}_{\text{suffix}}(t_{\text{crit}},\tau),\\
\varepsilon_{low}, & \text{else},
\end{cases}
\label{eq:elc-epslow}
\end{equation}
where $\mathcal{C}_{\text{crit}}$ is the set of first irrecoverable steps returned by
Algorithm~\ref{alg:binary-error-localization}, and $t_{\text{crit}}\in\mathcal{C}_{\text{crit}}$ denotes the localized critical step on trajectory $\tau$.
$\mathcal{I}_{\text{suffix}}(t_{\text{crit}},\tau)$ denotes the set of agent-generated token positions in $\tau$ from step $t_{\text{crit}}$ to the end of the trajectory, namely all tokens in $\{\bm{a}_t\}_{t=t_{\text{crit}}}^{T}$.
To see the effect, consider the clipped term $\min(\rho A,\ \mathrm{clip}(\rho,1-\varepsilon_{\text{low}}^{t},1+\varepsilon_{\text{high}})A)$.
When $A<0$, decreasing $1-\varepsilon_{\text{low}}^{t}$ allows smaller ratios $\rho<1$ to remain unclipped, which increases the magnitude of the negative update and penalizes error-inducing suffix tokens more aggressively.
For $A>0$, we keep the upper bound $1+\varepsilon_{\text{high}}$ unchanged, so the effect is limited mainly to discouraging the error suffix rather than amplifying positive updates.
The hyperparameter $\varepsilon_{\text{elc}}\ge 0$ controls the relaxation degree, enabling stronger corrections on the error-related suffix while preserving the training stability.

\subsection{Training objective of ELPO}
ELPO follows the GRPO training objective in Eq.~\eqref{eq:grpo-obj}, but modifies the token-level surrogate in two places:
we use the hierarchical advantage $A^{\text{hier}}$ in place of the group-level advantage $\hat{A}_i$, and we replace the constant lower clipping bound $1-\varepsilon_{low}$ with the token-dependent bound $1-\varepsilon_{\text{low}}^{t}$ in Eq.~\eqref{eq:elc-epslow}.
This yields the ELPO clipped surrogate loss
\begin{equation}
\label{eq:elpo-clip}
\small{
\begin{aligned}
L_{\text{CLIP}}^{\text{ELPO}}(\theta,i,t)
  &= \min\Big(
      \rho_{i,t}(\theta)A^{\text{hier}},\; \\[2pt]
  &\qquad
      \mathrm{clip}\big(
        \rho_{i,t}(\theta), 1-\varepsilon_{\text{low}}^{t}, 1+\varepsilon_{high}
      \big)A^{\text{hier}}
    \Big)
\end{aligned}
}
\end{equation}
which is plugged into Eq.~\eqref{eq:grpo-obj} with the same masking scheme.

\begin{table*}[t]
\centering
\footnotesize
\caption{Overall performance on TIR benchmarks (higher is better, \%), grouped by domain. 
\textbf{From} denotes the starting checkpoint type (\textsc{Base}/\textsc{Inst}/\textsc{Math-Inst}). 
RL results are grouped by the backbone family used for training (Qwen2.5-7B vs.\ Qwen3-4B). 
The top two results of different baseline categories in each column are \textbf{bolded} and \underline{underlined}.}
\vspace{-0.8em}
\label{tab:overall_results}
\setlength{\tabcolsep}{3.6pt}
\renewcommand{\arraystretch}{1.05}

\resizebox{\textwidth}{!}{%
\begin{tabular}{p{2.4cm} p{1.3cm} ccccc cc c}
\toprule
\multirow{2}{*}{\textbf{Method}} &
\multirow{2}{*}{\textbf{From}} &
\multicolumn{5}{c}{\textbf{Mathematical Reasoning}} &
\multicolumn{1}{c}{\textbf{Science}} &
\multicolumn{1}{c}{\textbf{Code}} &
\multirow{2}{*}{\textbf{Avg.}} \\
\cmidrule(lr){3-7} \cmidrule(lr){8-8} \cmidrule(lr){9-9}
& & AIME24 & AIME25 & MATH500 & GSM8K & MATH &
\shortstack{GPQA-Diamond} &
\shortstack{LiveCodeBench} & \\
\midrule

\multicolumn{10}{c}{\textit{\textbf{Prompting / Inference Baselines}}} \\
\midrule

\multicolumn{10}{l}{\textit{\textbf{Self-Contained Reasoning}}} \\
Qwen2.5-7B & Inst & 16.7 & 10.0 & 72.2 & \underline{91.6} & 75.5 & 31.3 & 15.2 & 44.6 \\
Qwen3-4B   & Inst & \textbf{63.3} & \textbf{47.4} & \underline{84.8} & \textbf{92.0} & \underline{80.4} & \underline{52.0} & \textbf{35.1} & \textbf{66.4} \\
DeepSeek-V3& Base & \underline{39.2} & \underline{28.8} & \textbf{90.2} & 89.3 & 61.6  & \textbf{59.1} & 16.1 & \underline{54.9} \\

\addlinespace[0.35em]
\hdashline
\addlinespace[0.35em]

\multicolumn{10}{l}{\textit{\textbf{TIR Reasoning}}} \\
Qwen2.5-7B & Inst & 6.7 & 10.0 & 68.2 & 64.6 & 78.2 & 25.5 & 12.2 & 37.9 \\
Qwen3-4B   & Inst & 17.0 & 16.3 & 73.0 & 88.2 & \textbf{81.5} & 44.3 & \underline{23.0} & 49.0 \\

\midrule
\multicolumn{10}{c}{\textit{\textbf{RL Methods based on Qwen2.5-7B}}} \\
\midrule
\multicolumn{10}{l}{\textit{\textbf{Classical RL}}} \\
GRPO        & Inst & 23.3 & 26.7 & 78.0 & \textbf{92.8} & 87.8 & 32.7 & 16.8 & 51.2 \\
Reinforce++ & Instruct & 26.7 & 23.3 & 78.8 & 92.2 & 88.8 & 31.3 & 17.1 & 51.2 \\
\addlinespace[0.35em]
\hdashline
\addlinespace[0.35em]

\multicolumn{10}{l}{\textit{\textbf{Clipping-optimized RL}}} \\
DAPO & Instruct & 20.0 & 23.3 & 80.4 & 91.0 & 88.8 & 33.5 & 18.5 & 50.8 \\
\addlinespace[0.35em]
\hdashline
\addlinespace[0.35em]
\multicolumn{10}{l}{\textit{\textbf{Agentic RL}}} \\
ToRL      & Math-Inst & 40.2 & 27.9 & 82.2 & 91.4 & 87.8 & 51.5 & 17.8 & 57.0 \\
ARPO      & Inst & 30.0 & 30.0 & 78.8 & 92.2 & 88.8 & \textbf{53.0} & 18.3 & 55.9 \\
AEPO      & Inst & 33.3 & 30.0 & 80.4 & 92.2 & 90.0 & \textbf{53.0} & 17.8 & 56.7 \\
CIR       & Math & 42.3 & 29.2 & 86.4 & 91.8 & 90.4 & 49.5 & \underline{19.3} & 58.4 \\
GIGPO     & Inst & 30.0 & 20.0 & 78.4 & 91.6 & 87.6 & 52.5 & 18.1 & 54.0 \\
DemyAgent & Inst & \underline{43.2} & \underline{30.5} & \underline{87.9} & \underline{92.6} & \underline{90.8} & 52.0 & 18.9 & \underline{59.4} \\
\rowcolor[HTML]{E8F3FF}
ELPO-7B (Ours) & Inst & \textbf{48.1} & \textbf{31.1} & \textbf{89.4} & \textbf{92.8} & \textbf{91.2} & \underline{52.7} & \textbf{19.7} & \textbf{60.7} \\
\midrule

\multicolumn{10}{c}{\textit{\textbf{RL Methods based on Qwen3-4B}}} \\
\midrule
\multicolumn{10}{l}{\textit{\textbf{Additional Comparison}}} \\
DemyAgent-4B & Inst & \underline{72.6} & \textbf{70.0} & \textbf{90.4} & \underline{93.1} & \underline{91.6} & \underline{58.5} & \underline{26.8} & \underline{71.9} \\
\rowcolor[HTML]{E8F3FF}
ELPO-4B (Ours) & Inst & \textbf{74.3} & \underline{69.4} & \underline{89.6} & \textbf{94.5} & \textbf{92.8} & \textbf{59.1} & \textbf{28.6} & \textbf{72.6} \\
\bottomrule
\end{tabular}%
}
\vspace{-0.6em}
\end{table*}

\section{Experimental Setup}
\subsection{Datasets}
We evaluate ELPO across three long-horizon TIR reasoning tasks:
\begin{enumerate}[leftmargin=*, itemsep=1pt, topsep=2pt, parsep=0pt, partopsep=0pt]
  \item \textbf{Mathematical Reasoning.} This includes MATH~\cite{MATH}, GSM8K~\cite{GSM8K}, and competition-level challenges such as MATH500~\cite{MATH}, AIME2024\footnote{\url{https://huggingface.co/datasets/Maxwell-Jia/AIME_2024}}, and AIME2025\footnote{\url{https://huggingface.co/datasets/opencompass/AIME2025}}.
  \item \textbf{Knowledge-Intensive Scientific Reasoning.} We adopt GPQA-Diamond~\cite{GPQA-Diamond} as a challenging science QA benchmark that requires multi-step factual reasoning and verification.
  \item \textbf{Code Generation and Execution.} We use LiveCodeBench-v6~\cite{livecodebench} to assess executable code synthesis, where solutions may require iterative tool-based execution.
\end{enumerate}
All dataset splits align with the standard settings established by previous works~\cite{ARPO, AEPO, DemyAgent}.

\subsection{Baselines}
We consider the following strong baselines:
\begin{enumerate}[leftmargin=*, label=(\arabic*), itemsep=1pt, topsep=2pt, parsep=0pt, partopsep=0pt]
  \item \textbf{Advanced RL Algorithms.} We compare ELPO with three categories of RL methods commonly used for LLM reasoning: (i) Vanilla RL: GRPO~\cite{GRPO} and Reinforce++~\cite{Reinforce++}; (ii) Clipping-optimized RL: DAPO~\cite{DAPO}; (iii) Agentic RL: ToRL~\cite{ToRL}, ARPO~\cite{ARPO}, AEPO~\cite{AEPO}, CIR~\cite{CIR}, GIGPO~\cite{GIGPO}, and DemyAgent~\cite{DemyAgent}.
  \item \textbf{Advanced Backbone Models.} We compare ELPO with advanced backbones, including Qwen2.5~\cite{Qwen2.5} and Qwen3~\cite{Qwen3}, as well as strong open-source reasoning models such as DeepSeek-V3~\cite{Deepseek-V3} and DeepSeek-R1.
\end{enumerate}

\subsection{Implementation Details}
We implement ELPO on top of the VERL framework\footnote{\url{https://github.com/volcengine/verl}} with feedback-token masking, excluding tool-call feedback tokens from the loss so gradients are accumulated only over agent-generated tokens.
We conduct all training on two backbone instruction-tuned LLMs, Qwen2.5-7B-Instruct\footnote{\url{https://huggingface.co/Qwen/Qwen2.5-7B-Instruct}} and Qwen3-4B-Instruct\footnote{\url{https://huggingface.co/Qwen/Qwen3-4B-Instruct-2507}}.
Following DemyAgent~\cite{DemyAgent}, we first perform a cold-start SFT for $5$ epochs on the Open-AgentRL-SFT-3K dataset\footnote{\url{https://huggingface.co/datasets/Gen-Verse/Open-AgentRL-SFT-3K}} to acquire basic tool-use patterns, and then run RL training for $1$ epoch on the Open-AgentRL-30K dataset\footnote{\url{https://huggingface.co/datasets/Gen-Verse/Open-AgentRL-30K}} under the same data construction protocol.
The GRPO KL regularization coefficient is set to $0$ for stability.

Our training uses a total batch size of $128$ with a PPO mini-batch size of $16$, and a maximum context length of $20\text{K}$ tokens.
For every method, we enforce the same per-input rollout budget $N_{\text{total}}=16$.
ELPO uses no extra samples: BEL reuses this fixed budget by reallocating rollouts between initial full trajectories and anchor-prefix suffix branches (all counted within $N_{\text{total}}$).
All baselines are run under the same $N_{\text{total}}$ budget; for tree-based baselines, every branch expansion is also counted as one rollout within $N_{\text{total}}$.
All experiments are conducted on $8$ NVIDIA A100 GPUs.
Details hyperparameter settings for ELPO method are provided in Appendix~\ref{appendix:ELPO hyperparameter}.
\section{Experimental Results}
\subsection{Main Result on Different TIR Tasks}
To validate the effectiveness of ELPO for TIR, we compare it with prompting baselines and representative RL algorithms across mathematical reasoning, science QA, and code execution.
Table~\ref{tab:overall_results} summarizes the results. We highlight three observations.

\noindent\textbf{(1) Naive tool-enabled prompting can degrade performance.}
Simply enabling tool calls at inference time does not guarantee gains.
On Qwen2.5-7B, TIR prompting underperforms self-contained prompting on average, with pronounced drops on difficult math and GSM8K.
This suggests that multi-step tool interaction increases error propagation and makes failures harder to recover without learning-based adaptation.

\noindent\textbf{(2) RL substantially improves TIR reasoning capabilities over prompting baselines.}
Training with outcome-based RL consistently improves over TIR prompting and yields strong performance across domains.
Notably, Agentic RL baselines outperform classical trajectory-level RL on average, supporting the benefit of introducing step-aware mechanisms when supervision is only available at the terminal outcome.

\noindent\textbf{(3) ELPO delivers the best overall results and scales to stronger backbones.}
ELPO achieves the highest average score on both Qwen2.5-7B and Qwen3-4B.
Compared with the strongest prior Agentic RL baseline DemyAgent, ELPO improves the average accuracy by 2.2\% on Qwen2.5-7B and by 1.0\% on Qwen3-4B.
These gains are consistent with ELPO’s design, which localizes the earliest error-inducing decision via binary-search tree rollouts and derives fine-grained advantages by contrasting branches while retaining trajectory-level returns.

\begin{figure}[t]
  \centering
  \includegraphics[width=\columnwidth]{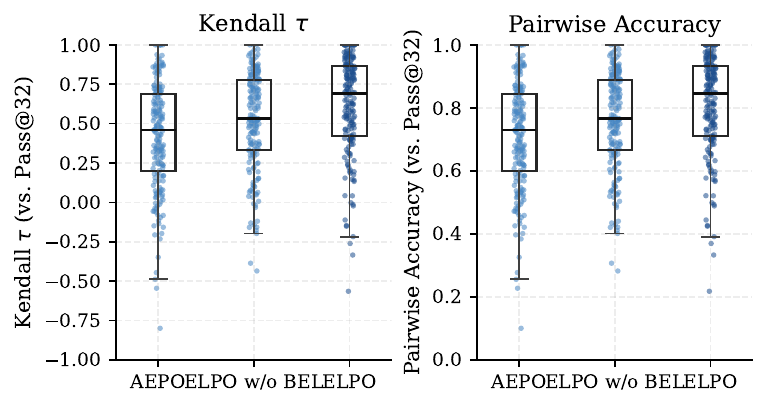}
  \caption{Local ranking quality at branching prefixes (pairwise accuracy / Kendall’s $\tau$ vs Mean@32 reference).}
  \label{fig:ranking}
\end{figure}

\subsection{Ranking Quality Analysis in Rollouts}
In Agentic RL, the learning signal ultimately depends on whether the algorithm can reliably distinguish which rollout candidates are better. 
Compared to linear rollouts that rely primarily on outcome rewards, recent tree rollout methods~\cite{ARPO,AEPO,Tree-GRPO} explicitly introduce a branching structure, which strengthens the algorithm's ability to rank alternative answers and intermediate decisions.
We therefore conduct a focused evaluation of the ranking quality induced by tree rollout methods including ELPO.

We randomly sample 200 questions from the DemyAgent training set covering a range of difficulty levels and perform rollouts with the Qwen2.5-7B-Instruct model.
To isolate the contribution of BEL in tree construction, we additionally evaluate a variant that removes BEL and instead selects branching prefixes uniformly at random, following the TreeGRPO-style~\cite{Tree-GRPO} rollout procedure, which is denoted as ELPO w/o BEL.
After constructing the rollout tree for each method, we collect all \emph{branching prefixes} (i.e., prefixes with multiple child nodes).
At each branching prefix, we induce a local ordering over sibling branches using the method’s own step-level branch scores on the child nodes (e.g., ELPO’s hierarchical advantage; Section~\ref{HAC}).

To obtain a reference ranking, we approximate each child branch’s quality via Monte Carlo estimation with $K{=}32$ samples.
For each branching prefix and each of its child prefixes, we draw $K{=}32$ rollouts from that child prefix and compute the success rate (mean binary correctness), denoted as Mean@32, which induces a reference ordering among sibling branches.
We then compare each method-induced ordering against this Mean@32 reference and quantify ranking quality using pairwise accuracy and Kendall's $\tau$~\cite{Kendall}.

As shown in Figure~\ref{fig:ranking}, ELPO achieves the highest agreement with the reference ranking, while ELPO w/o BEL still consistently outperforms AEPO. 
This aligns with our design: (i) ELPO's advantage modeling is explicitly tailored to local preference comparisons at branching points, producing more comparable signals among sibling branches under the same parent; and (ii) BEL concentrates expansion budget around decision steps that are more likely to induce meaningful divergence and reveal failure causes, improving the informativeness of sibling-branch comparisons and thus further strengthening local ranking accuracy.

More detailed experimental results are provided in the appendix, including ELPO-specific hyperparameter settings in Appendix~\ref{appendix:ELPO hyperparameter}, Pass@K and Major@K sampling analysis in Appendix~\ref{appendix:pass@k}, consistency evaluation of localizing the first irrecoverable step in Appendix~\ref{appendix:consistency}, component ablations in Appendix~\ref{appendix:ablation}, hyperparameter sensitivity analysis in Appendix~\ref{appendix:sensitive}, training dynamics in Appendix~\ref{appendix:trainingdynamics}, and qualitative case studies in Appendix~\ref{appendix:case}.

\section{Conclusion}
We study outcome-only Agentic RL for long-horizon TIR, where sparse terminal rewards lead to diffuse credit assignment and obscure the earliest irrecoverable decision. We propose \textbf{Error-Localized Policy Optimization (ELPO)}, which uses binary-search tree rollouts to localize the \emph{first irrecoverable step} under a fixed rollout budget, and turns this signal into stable updates via hierarchical advantage attribution and error-localized adaptive clipping. Experiments on tool-augmented benchmarks in math, science QA, and code execution show that ELPO consistently outperforms strong Agentic RL baselines under comparable sampling budgets, with better Pass@K/Major@K scaling and improved tool-call efficiency. Overall, ELPO demonstrates that explicitly localizing and exploiting irrecoverable errors provides a practical path to more targeted and reliable optimization for long-horizon agentic reasoning.

\section*{Limitations}
Despite its effectiveness, ELPO has several limitations. 
First, the localization of the first irrecoverable step is an empirical result obtained under a fixed rollout budget, current policy, and decoding configuration, based on finite suffix sampling. In low-success or highly stochastic settings, rare successes or false negatives may introduce bias or variance into the binary localization process, and the identified step may shift as the policy or sampling budget changes. 
Second, our evaluation focuses on relatively deterministic tool environments such as code execution, calculation, and closed-form question answering, and we have not systematically tested ELPO in noisier or more interactive settings, e.g., open-domain web search or GUI-based tools, where tool feedback uncertainty may further affect recoverability judgments. 
Finally, ELPO currently targets a single earliest critical step for localized updates, which may not fully capture failure modes arising from multiple interacting errors or gradual error accumulation.

\section*{Ethical Considerations}
This work improves reinforcement learning for TIR by enabling finer-grained error localization and credit assignment under a fixed rollout budget. We follow standard academic ethical practices and use only publicly available tools and benchmark environments (e.g., code execution and calculators), without relying on private data or undisclosed tool access. Although stronger tool-use ability could be misused in deployment, ELPO is a training technique rather than a mechanism for data collection or deception; responsible use should include appropriate access control and auditing. We do not anticipate direct harm to individuals or groups from the experiments reported in this paper.
\bibliography{custom}

\appendix

\section{Pseudo Code of ELPO's Rollout Strategy}
\label{appendix:Pseudo Code}
The pseudo-code of the graph construction process
is shown in Algorithm~\ref{alg:binary-error-localization}.

\begin{algorithm}
\caption{Tree Rollout via Binary Error Localization}
\label{alg:binary-error-localization}
\footnotesize
\begin{algorithmic}[1]
\Require task input $x$, tool set $\mathcal{T}$, old policy $\pi_{\theta_{\text{old}}}$, rollout budget $N_{\text{total}}$, buffer size $B_{\max}$, min/max branch counts per anchor $(X_{\min},X_{\max})$, entropy scaling factor $\beta$
\State Initialize: $\mathcal{B}\gets\emptyset$, $\mathcal{G}_x\gets\emptyset$, $\mathcal{C}_{\text{crit}} \gets \emptyset$, rollouts $\gets 0$
\State \textcolor{blue}{// Main loop: repeat until rollout budget is exhausted}
\While{rollouts $< N_{\text{total}}$}
    \State \textcolor{blue}{// Collect error trajectories}
    \While{$|\mathcal{B}| < B_{\max}$ \textbf{and} rollouts $< N_{\text{total}}$}
        \State sample a full trajectory  $\tau$ from $\pi_{\theta_{\text{old}}}$ with tools $\mathcal{T}$
        \State rollouts $\gets$ rollouts + 1
        \State compute terminal reward $R(\tau)$
        \If{$R(\tau)=0$}
            \State compute $H^{\text{root}}(\tau)$ and $H^{\text{tool}}(\tau)$
            \State $\mathcal{B}\gets \mathcal{B}\cup\{(\tau, H^{\text{root}}(\tau), H^{\text{tool}}(\tau))\}$
        \EndIf
    \EndWhile
    \If{$|\mathcal{B}| = 0$}
        \State \textbf{break}
    \EndIf
    \State $\tau_{\text{search}}^{*}
    = \arg\max_{\tau_i \in \mathcal{B}} \big( H^{\text{tool}}(\tau_i) - H^{\text{root}}(\tau_i) \big),$
    \State $\mathcal{B}\gets\emptyset$
    \State add $\tau_{\text{search}}^{*}$ as the reference path into $\mathcal{G}_x$
    \State \textcolor{blue}{// Binary tree-search over anchor steps on $\tau_{\text{search}}^{*}$}
    \State let $K$ be the number of tool-call steps in $\tau_{\text{search}}^{*}$
    \State set $L\gets 1$, $R\gets K$
    \While{$L<R$ \textbf{and} rollouts $< N_{\text{total}}$}
        \State $m\gets \lfloor (L+R)/2 \rfloor$ \Comment{\textcolor{blue}{// anchor step}}
        \State take $\text{prefix}(m)$ and compute $H^{\text{tool}}_m$
        \State $X_m \gets X_{\min} + (X_{\max}-X_{\min})\cdot \sigma\!\big(\beta(H^{\text{tool}}_m - H^{\text{root}}(\tau_{\text{search}}^{*}))\big)$
        \State $\text{reachable}\gets\textbf{false}$
        \For{$k=1$ \textbf{to} $X_m$}
            \If{rollouts $\ge N_{\text{total}}$} \State \textbf{break} \EndIf
            \State sample one trajectory $\hat{\tau}^{(k)}$ from $\text{prefix}(m)$
            \State rollouts $\gets$ rollouts + 1
            \State add $\hat{\tau}^{(k)}$ as a child branch at step $m$ in $\mathcal{G}_x$
            \State compute terminal reward $R(\hat{\tau}^{(k)})$
            \If{$R(\hat{\tau}^{(k)})=1$}
                \State $\text{reachable}\gets\textbf{true}$
            \EndIf
        \EndFor
        \If{$\text{reachable}$}
            \State $L\gets m+1$ \Comment{\textcolor{blue}{// error likely after $m$}}
        \Else
            \State $R\gets m$ \Comment{\textcolor{blue}{// error likely at or before $m$}}
        \EndIf
    \EndWhile
    \State $t_{\text{crit}}\gets L$
    \State $\mathcal{C}_{\text{crit}} \gets \mathcal{C}_{\text{crit}} \cup \{t_{\text{crit}}\}$
\EndWhile
\State \textbf{Output}: error-localized rollout tree $\mathcal{G}_x=(\mathcal{S},\mathcal{E})$, first irrecoverable steps $\mathcal{C}_{\text{crit}}$
\end{algorithmic}
\end{algorithm}

\begin{figure*}[t]
  \centering
  \includegraphics[width=\textwidth]{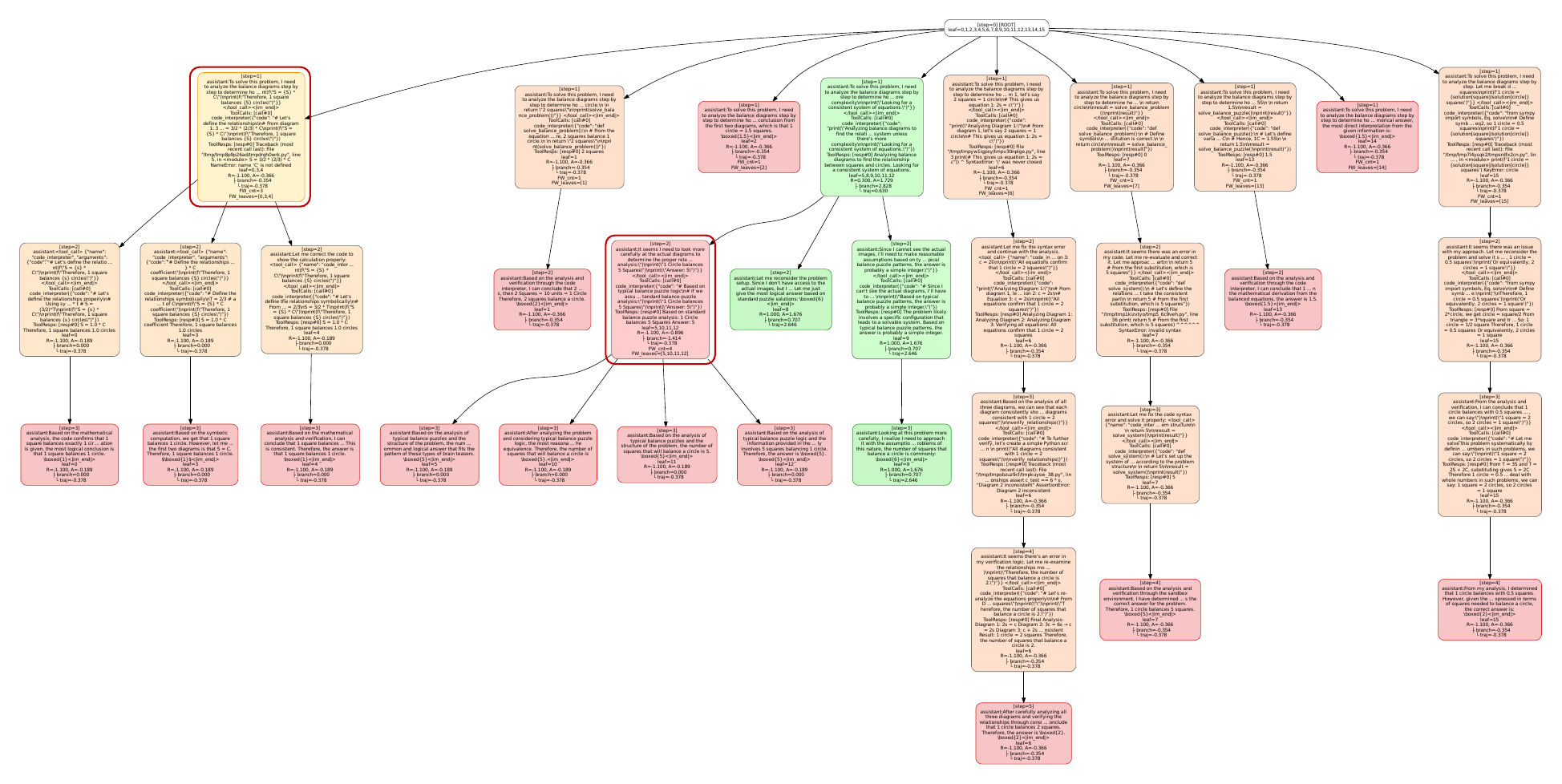}
  \caption{An example of a rollout tree based on BEL. The red box outlines the first irrecoverable step identified by the rollout tree.}
  \label{fig:example of bel tree}
\end{figure*}

\section{An example of a BEL rollout tree}
\label{example if BEL tree}
Figure~\ref{fig:example of bel tree} provides an example of the rollout tree for a certain mathematical problem when the rollout budget $N$ is set to 16.

\section{ELPO-related Hyperparameter Settings}
\label{appendix:ELPO hyperparameter}
\begin{table}[t]
\centering
\small
\setlength{\tabcolsep}{6pt}
\renewcommand{\arraystretch}{1.15}
\begin{tabular}{l l l}
\toprule
\textbf{Module} & \textbf{Hyperparameter} & \textbf{Value} \\
\midrule
Tree rollout
& $N_{\text{total}}$ & 16\\
& $B_{\max}$ & 3\\
& $X_{\min}$ & 1\\
& $X_{\max}$ & 3\\
& $\beta$ & 5\\

\midrule
Advantage attribution
& $\lambda_{\text{tree}}$ & 0.5\\

\midrule
ELC
& $\varepsilon_{\text{low}}$ & 0.2\\
& $\varepsilon_{\text{high}}$ & 0.315\\
& $\varepsilon_{\text{elc}}$ & 0.115\\
\bottomrule
\end{tabular}
\caption{ELPO-specific hyperparameters used in our implementation.}
\label{tab:elpo-hparams}
\end{table}
We report the ELPO-specific hyperparameters used in our implementation in
Table~\ref{tab:elpo-hparams}.

\section{Pass@K and Major@K Sampling Analysis}
\label{appendix:pass@k}
Due to the stochasticity in tool-integrated rollouts, single-sample evaluation can underestimate an agent’s true capability.
We therefore report Pass@K and Major@K with $K\in\{4,16,32\}$.

\begin{figure*}[t]
  \centering
  \includegraphics[width=0.85\textwidth]{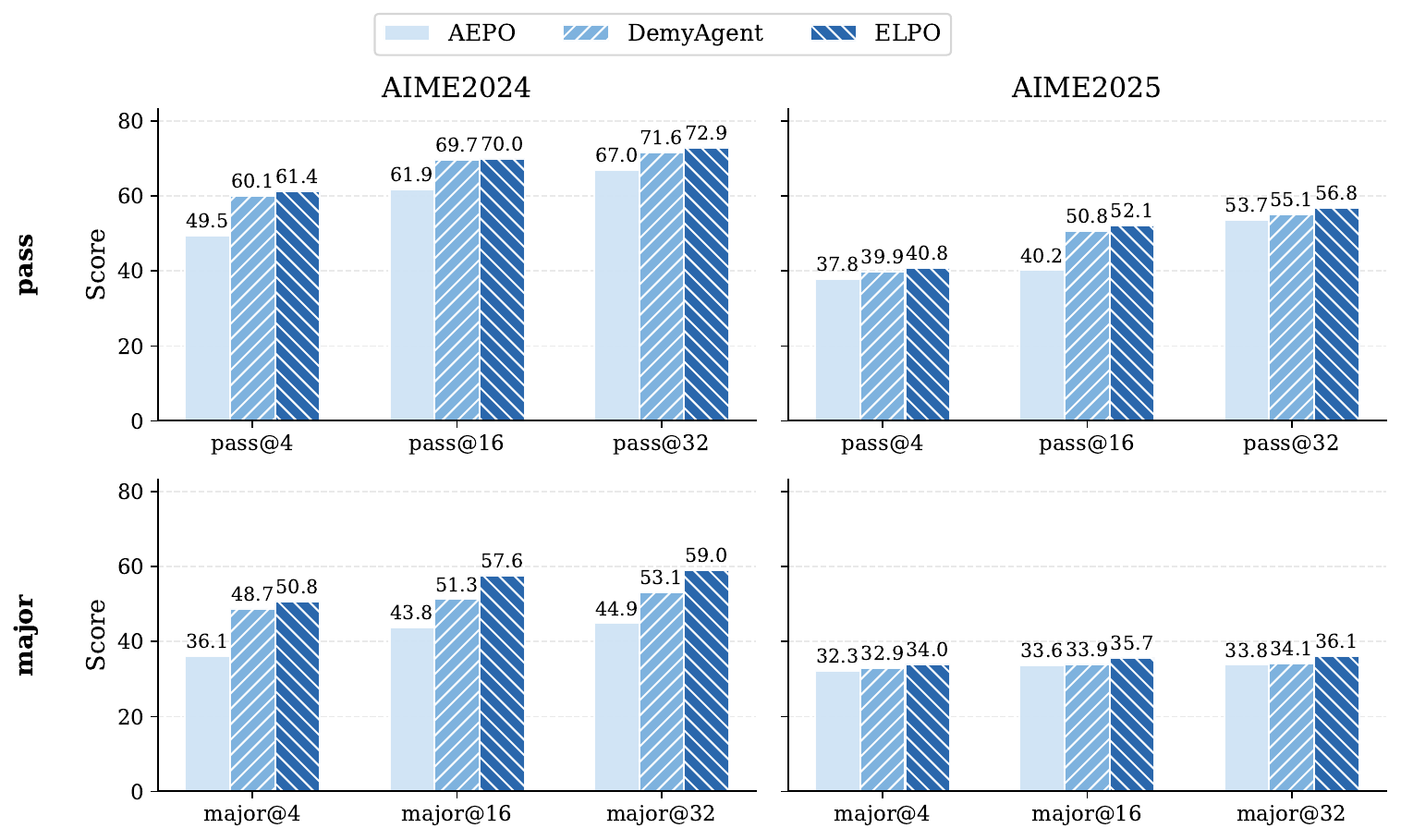}
  \caption{Pass@K and Major@K sampling analysis on AIME2024/2025.}
  \label{fig:pass@k}
\end{figure*}

As shown in Figure~\ref{fig:pass@k}, ELPO consistently benefits more from larger-scale sampling and achieves the best scores across settings.
On AIME2024 and AIME2025, ELPO consistently outperforms both AEPO and DemyAgent, with the improvements most pronounced on the Major@K metric (especially at larger $K$).
We attribute the stronger scaling on Major@K to ELPO’s localized supervision.
BEL concentrates branching near the earliest decisive error step, yielding more informative sibling-branch comparisons, while our hierarchical advantage produces directly comparable preference signals within each branching point.
Together, this improves both sampling success (Pass@K) and sample-level consensus (Major@K) as $K$ grows.

\section{Consistency of First Irrecoverable Step Localization}
\label{appendix:consistency}

To validate that ELPO’s BEL rollout mechanism truly pinpoints the \emph{first irrecoverable step}, we construct an objective ground truth based on Pass@K recoverability tests on failed traces.
We run ELPO training from Qwen2.5-7B-Instruct on AIME2024.
At training step 200, we construct BEL rollout trees for the sampled questions.
For each instance whose rollout fails, we record the critical step index $t_{\text{crit}}$ localized by BEL.

For each failed trajectory with $K$ decision/tool-call steps, we compute a recoverability indicator for every step $t \in \{1,\ldots,K\}$ by freezing the prefix up to step $t$ and resampling $k_{\text{eval}}$ suffix completions under the same tool environment.
A prefix is marked \emph{recoverable} if any suffix completion succeeds (i.e., reaches Pass@$k_{\text{eval}}{=}1$), and \emph{irrecoverable} otherwise.
We then define the ground-truth first irrecoverable step as the earliest irrecoverable step.
We measure exact-match Hit@1, i.e., the fraction of failed trajectories where $t_{\text{crit}}$ exactly matches $t^{*}$.

Table~\ref{tab:consistency_main} shows that ELPO achieves substantially higher agreement than random selection and an entropy-peak heuristic, supporting that BEL reliably identifies the earliest irrecoverable decision that dominates failure under a fixed sampling budget.

\begin{table}[t]
\centering
\small
\setlength{\tabcolsep}{6pt}
\renewcommand{\arraystretch}{1.1}
\caption{Agreement on localizing the first irrecoverable step in failed AIME2024 traces.
Random step uniformly samples a step index; Entropy-peak picks the step with the maximum tool-call entropy; ELPO localizes the step via BEL.}
\label{tab:consistency_main}
\begin{tabular}{l c}
\toprule
Method & Hit@1 \\
\midrule
Random step & 32.6\% \\
Entropy-peak step & 47.4\% \\
ELPO & 66.8\% \\
\bottomrule
\end{tabular}
\end{table}

\section{Ablation Study on ELPO Components}
\label{appendix:ablation}

\begin{table}[t]
\centering
\small
\setlength{\tabcolsep}{7pt}
\renewcommand{\arraystretch}{1.15}
\caption{
Ablation study of ELPO under the same rollout budget ($N_{total}=16$) for all methods.
We report accuracy (Mean@32) on AIME2024 and AIME2025. The top two results in each column are \textbf{bolded} and \underline{underlined}.
}
\label{tab:ablation_aime_singlecol}
\begin{tabular}{l cc}
\toprule
\textbf{Variant} & \textbf{AIME24} & \textbf{AIME25} \\
\midrule
\rowcolor[HTML]{E8F3FF}
\textbf{ELPO} & \textbf{48.1} & \textbf{31.1} \\
\midrule
w/o BEL & 45.8 & 29.1 \\
w/o FAA & 43.4 & 28.9 \\
w/o ELC & \underline{47.2} & 29.8 \\
w/o entropy-gap selection & 46.0 & \underline{30.6} \\
w/o adaptive $X_m$ & 46.7 & 30.4 \\
\bottomrule
\end{tabular}
\end{table}

Table~\ref{tab:ablation_aime_singlecol} reports ELPO component ablations under the same rollout budget ($N_{total}=16$), evaluated by Mean@32 on AIME2024 and AIME2025. Removing BEL consistently degrades performance, showing that localizing the first irrecoverable step is crucial under sparse terminal rewards. Disabling FAA causes an even larger drop, suggesting ELPO’s gain comes from turning the localized tree into actionable fine-grained credit rather than merely “using a tree”. Within BEL, replacing entropy-gap trajectory selection with uniform sampling or replacing adaptive suffix allocation $X_m$ with a fixed branch count both hurt performance, indicating these heuristics improve localization robustness and sibling-branch informativeness under tight budgets. Finally, removing ELC slightly improves AIME2024 but hurts AIME2025, reflecting a stability--aggressiveness trade-off of relaxing clipping around the localized critical-step suffix.

\section{Sensitivity Analysis of ELPO Hyperparameters}
\label{appendix:sensitive}
\begin{figure}[t]
  \centering
  \includegraphics[width=\columnwidth]{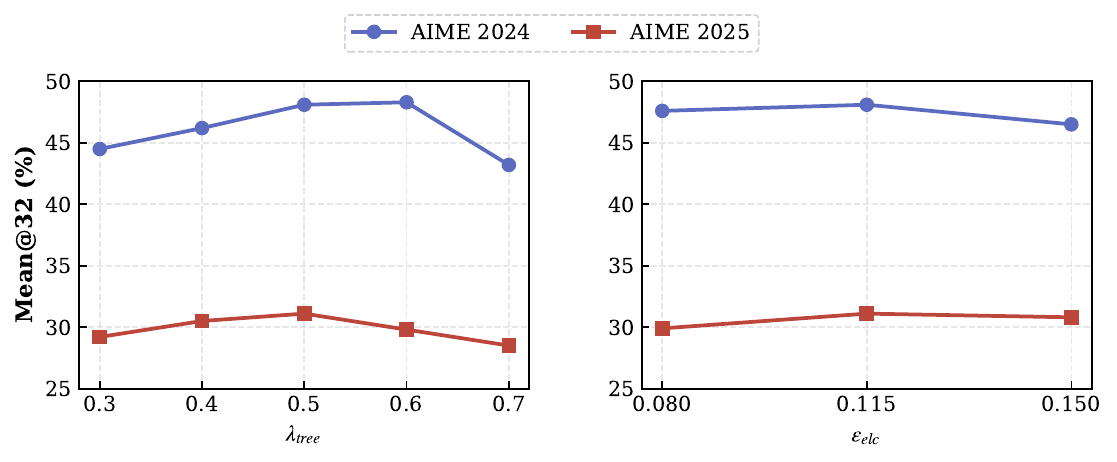}
  \caption{Sensitivity of ELPO performance (Mean@32) to the hierarchical advantage weight $\lambda_{\text{tree}}$ (left) and the error-localized clipping parameter $\varepsilon_{\text{elc}}$ (right) on AIME 2024 and AIME 2025.}
  \label{fig:sensitivity}
\end{figure}

Figure~\ref{fig:sensitivity} presents the sensitivity of ELPO to key hyperparameters on AIME 2024 and AIME 2025.
For the hierarchical advantage weight $\lambda_{\text{tree}}$, performance peaks at intermediate values, while overly large $\lambda_{\text{tree}}$ leads to a clear degradation in Mean@32 on both benchmarks.
This suggests that relying too heavily on branch-level preference signals is insufficient for stable optimization, as it weakens the global constraint provided by trajectory-level advantages and biases updates toward local comparisons.
As a result, local preference signals may be over-emphasized in long-horizon decision-making, affecting overall solution quality.
In contrast, intermediate values of $\lambda_{\text{tree}}$ achieve a better balance between local discriminability and global consistency.

For the error-localized clipping parameter $\varepsilon_{\text{elc}}$, performance varies more moderately within the tested range, with intermediate values generally yielding the best results.
When $\varepsilon_{\text{elc}}$ is too small, the corrective update on critical steps is limited, whereas overly large values can lead to overly aggressive updates and reduced training stability.
Overall, ELPO exhibits robust performance across a reasonable range of hyperparameter values and achieves optimal or near-optimal results with intermediate settings.

\section{Training Dynamics}
\label{appendix:trainingdynamics}
\begin{figure*}
\centering
  \includegraphics[width=1\textwidth]{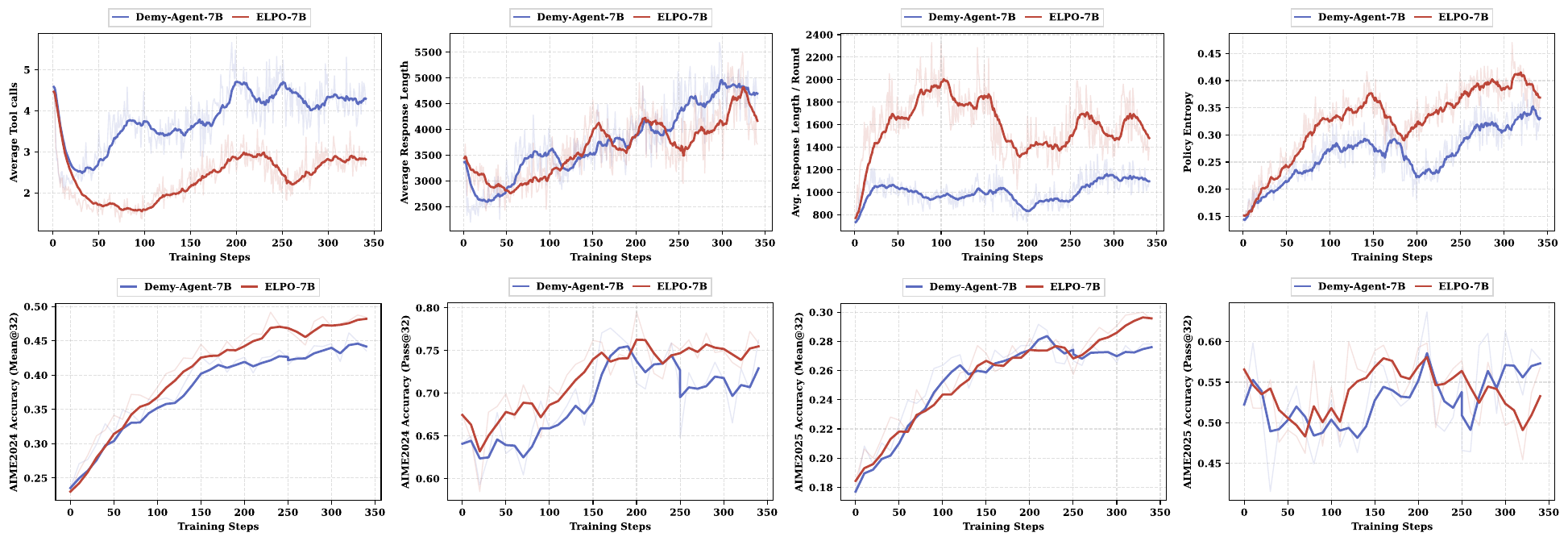}
  \caption{Training curves comparing ELPO and DemyAgent under the same training setup. We report tool usage, response length, policy entropy, and task performance (Mean@32 and Pass@32) on AIME2024 and AIME2025 over training.}
  \label{fig:curve}
\end{figure*}

Figure~\ref{fig:curve} compares ELPO with DemyAgent under the same backbone (Qwen2.5-7B-Instruct), SFT/RL data, and base training recipe, isolating the effect of the RL algorithm.
ELPO consistently reduces tool usage: tool calls drop faster and converge to a lower level (around three), while response length stays comparable to DemyAgent, indicating the gains are not from longer traces.
Instead, error-localized rollouts and credit assignment focus learning on the critical decision and its suffix, reducing redundant tool-invocation loops and making each interaction more informative.
As a result, ELPO achieves higher performance with fewer tool rounds, producing more compact trajectories without sacrificing content.
We also observe slightly higher and more stable policy entropy, suggesting localized optimization preserves exploration near the critical step while stabilizing elsewhere under sparse outcome-only supervision.

ELPO also delivers stronger training-time performance.
On both AIME2024 and AIME2025, it attains higher Mean@32 and Pass@32 for most of training, showing that error-localized supervision improves sampling success and high-budget performance.
Overall, the curves support our conclusion that ELPO improves solution quality with more efficient tool use and stable optimization dynamics.

\section{Case Study}
\label{appendix:case}

To contrast representative success and failure patterns of \textsc{ELPO} against strong baselines in long-horizon TIR, we further validate our method via case analysis.
Specifically, we select two challenging problems from AIME 2025 (Problem 15 and Problem 23) that the strong baseline DemyAgent-7B fails to solve, and compare the full solution trajectories produced by ELPO-7B and DemyAgent-7B (Figures~\ref{fig:part2}, \ref{fig:fail_case1}, \ref{fig:part2_trig_final}, and \ref{fig:fail_case2}).

Across both cases, DemyAgent-7B shows a consistent failure mode: once its initial strategy is blocked (e.g., brute-force timeouts or unreliable numerics), the trajectory quickly reaches the \emph{first irrecoverable step} and then repeats isomorphic but ineffective strategies; later, it may override tool feedback with unverified analytical intuition, leading to a ``tool--reasoning mismatch''.
In contrast, ELPO-7B exhibits a stable success pattern: on Problem 15, it abandons brute-force enumeration after observing timeouts and reformulates the task into a verifiable lower-complexity counting procedure; on Problem 23, it treats tool--reasoning conflicts as debugging signals and fixes issues such as double counting or boundary cases until tool verification and analytical reasoning agree.

These cases suggest that ELPO's gains come less from stronger mathematical ``intuition'' than from a more reliable long-horizon correction mechanism.
Under a fixed rollout budget, BEL pinpoints the first irrecoverable step so learning focuses on the truly failure-causing decision; FAA converts the rollout tree into actionable credit assignment by combining branch-level contrasts with trajectory-level ranking; and ELC enables stronger yet stable policy updates on the localized critical step and its suffix.
Together, these components reduce ``ineffective repetition plus tool-override hallucination,'' improving the stability and reliability of complex mathematical TIR.

\begin{figure*}[t!]
    \centering
    
    \begin{tcolorbox}[
        enhanced,
        title={\textbf{A Representative Answer by ELPO-7B on AIME 2025 Problem 15}},
        colframe=acl-blue, colback=white, coltitle=white, fonttitle=\bfseries,
        boxrule=1pt, drop shadow, width=\linewidth,
        bottomrule=0pt, sharp corners=south, arc=3mm 
    ]
        \begin{tcolorbox}[
            colback=gray!5, colframe=acl-blue, leftrule=3mm,
            title={\textbf{Problem}}, coltitle=white, fonttitle=\bfseries,
            arc=1mm, boxrule=0.5pt, left=2mm, right=2mm, top=2mm, bottom=2mm,
            fontupper=\small
        ]
            Analyze and solve the following math problem step by step. Let $N$ denote the number of ordered triples of positive integers $(a,b,c)$ such that $a,b,c\leq3^6$ and $a^3+b^3+c^3$ is a multiple of $3^7$. Find the remainder when $N$ is divided by $1000$.
            \par\vspace{0.1cm}
            \textbf{Ground Truth:} \colorbox{acl-green!20}{\textcolor{acl-green!60!black}{\textbf{735}}}
        \end{tcolorbox}

        \begin{thoughtblock}
            To solve this problem, I need to find the number of ordered triples $(a,b,c)$ where $a,b,c \leq 729$ and $a^3 + b^3 + c^3$ is divisible by $2187$. Direct enumeration ($729^3 \approx 3.8 \times 10^8$) would be computationally intensive. Let's write a Python script to enumerate all such triples.
        \end{thoughtblock}
        \begin{codeblock}
\begin{lstlisting}[style=normalcode]
count = 0
for a in range(1, 730):
    for b in range(1, 730):
        for c in range(1, 730):
            if (a**3 + b**3 + c**3) % 2187 == 0:
                count += 1
print(count)
\end{lstlisting}
        \end{codeblock}
        \begin{obsblock}
no stdout here (Timeout)
        \end{obsblock}
        
        \begin{thoughtblock}
            Let me simplify and just run a small part to make sure the logic works, then run the full computation in a more controlled way.
        \end{thoughtblock}
        \begin{codeblock}
\begin{lstlisting}[style=normalcode]
small_mod = 9 # 3^2
small_max = 3
count_small = 0
for a in range(1, small_max + 1):
    for b in range(1, small_max + 1):
        for c in range(1, small_max + 1):
            if (a**3 + b**3 + c**3) % small_mod == 0:
                count_small += 1
print(count_small)
\end{lstlisting}
        \end{codeblock}
        \begin{obsblock}
7
        \end{obsblock}

        \begin{thoughtblock}
            The small case gives 7. Now let's run the full computation. Let's take a different approach using mathematical insight (distribution of cubes).
        \end{thoughtblock}
        
        \vspace{0.1cm}
        \centering \small \color{gray} \textit{(Code execution continues on next page...)}
    \end{tcolorbox}

\end{figure*}

\begin{figure*}[t!]
    \centering
    
    \begin{tcolorbox}[
        enhanced,
        colframe=acl-blue, colback=white,
        boxrule=1pt, drop shadow, width=\linewidth,
        toprule=0pt, sharp corners=north, arc=3mm, top=5mm 
    ]
        \begin{codeblock}
\begin{lstlisting}[style=normalcode]
from collections import defaultdict
count = 0
mod = 3**7
max_val = 3**6
cube_mods = defaultdict(int)
# Precompute cube mod 2187
for i in range(1, max_val + 1):
    cube_mods[i**3 % mod] += 1
# Check sums (Still O(N^3) in structure)
for a in range(1, max_val + 1):
    for b in range(1, max_val + 1):
        for c in range(1, max_val + 1):
            if (cube_mods[a**3 % mod] + ... ) > 0: 
                if (a**3 + b**3 + c**3) % mod == 0:
                    count += 1
print(count)
\end{lstlisting}
        \end{codeblock}
        \begin{obsblock}
no stdout here (Timeout)
        \end{obsblock}

        \vspace{0.3cm}
        \begin{thoughtblock}[colback=insight-bg,borderline west={5pt}{0pt}{insight-border},title={\small \textcolor{insight-border}{\textbf{$\Rightarrow$ \ Correction \& Critical Insight}}},coltitle=insight-border]
            Let's take a completely different approach. I'll use a symmetry argument. For any fixed values of $a$ and $b$, we can solve for $c$: $c^3 \equiv -a^3 - b^3 \pmod{2187}$. So for each pair $(a,b)$, we need to count how many $c$ satisfy this equation. This reduces complexity significantly to $O(N^2)$.
        \end{thoughtblock}
        
        \begin{codeblock}
\begin{lstlisting}[style=normalcode]
count = 0
mod = 3**7
max_val = 3**6
targets = {}

# Precompute all cube residues frequencies
for c in range(1, max_val + 1):
    targets[c**3 % mod] = targets.get(c**3 % mod, 0) + 1

# Now count valid triples using O(N^2) loop
for a in range(1, max_val + 1):
    for b in range(1, max_val + 1):
        target_sum = (-a**3 - b**3) % mod
        if target_sum in targets:
            count += targets[target_sum]

print(count)
print(count % 1000)
\end{lstlisting}
        \end{codeblock}
        \begin{obsblock}
885735
735
        \end{obsblock}

        \begin{thoughtblock}
            The computation gives us a count of 885,735, and the remainder when divided by 1000 is 735.
        \end{thoughtblock}

        \begin{tcolorbox}[
            colback=acl-green!10, 
            colframe=acl-green!50, 
            width=8cm, halign=center, boxrule=0.5pt, arc=3mm, center,
            fontupper=\footnotesize 
        ]
            \textbf{Final Answer:} \boxed{735} \quad \textcolor{acl-green!60!black}{\large \checkmark} \ \textbf{\textcolor{acl-green!60!black}{Correct}}
        \end{tcolorbox}
        
    \end{tcolorbox}
    \caption{A Representative Answer by ELPO-7B on AIME 2025 Problem 15.}
    \label{fig:part2}
\end{figure*}

\begin{figure*}[t!]
    \centering
    
    \begin{tcolorbox}[
        enhanced,
        title={\textbf{A Representative Answer by DemyAgent-7B on AIME 2025 Problem 15}},
        colframe=fail-red, 
        colback=white, 
        coltitle=white, 
        fonttitle=\bfseries,
        boxrule=1pt, 
        drop shadow, 
        width=\linewidth,
        arc=3mm 
    ]
        \begin{tcolorbox}[
            colback=gray!5, colframe=fail-red, leftrule=3mm,
            title={\textbf{Problem}}, coltitle=white, fonttitle=\bfseries,
            arc=1mm, boxrule=0.5pt, left=2mm, right=2mm, top=1mm, bottom=1mm,
            fontupper=\small
        ]
            Let $N$ denote the number of ordered triples of positive integers $(a,b,c)$ such that $a,b,c\leq3^6$ and $a^3+b^3+c^3$ is a multiple of $3^7$. Find the remainder when $N$ is divided by $1000$.
        \end{tcolorbox}

        \begin{thoughtblock}
            To solve this problem, I need to find the number of ordered triples $(a,b,c)$ where each of $a$, $b$, $c$ is a positive integer less than or equal to $3^6 = 729$, and $a^3 + b^3 + c^3$ is divisible by $3^7 = 2187$. ... Let me write a Python program to compute this count.
        \end{thoughtblock}
        
        \begin{codeblock}
\begin{lstlisting}[style=normalcode]
def count_valid_triples():
    MOD = 3**7
    max_val = 3**6
    # Precompute all cubes modulo MOD for values from 1 to max_val
    cubes = [pow(i, 3, MOD) for i in range(1, max_val + 1)]
    count = 0
    # Check all ordered triples (a,b,c) where a,b,c <= max_val
    for a in range(1, max_val + 1):
        for b in range(1, max_val + 1):
            for c in range(1, max_val + 1):
                if (cubes[a-1] + cubes[b-1] + cubes[c-1]) % MOD == 0:
                    count += 1
    return count

result = count_valid_triples()
print(f"Total count: {result}")
print(f"Result modulo 1000: {result % 1000}")
\end{lstlisting}
        \end{codeblock}
        \begin{obsblock}
no stdout here
        \end{obsblock}
        
        \begin{tcolorbox}[
            enhanced,
            colback=fail-bg,       
            colframe=fail-red,     
            leftrule=5mm,          
            title={\small \textbf{$\triangle$ \ Repeated Failure Step}}, 
            coltitle=fail-red!80!black, 
            fonttitle=\bfseries,
            boxed title style={boxrule=0pt, colframe=white, colback=white},
            attach boxed title to top left={xshift=0mm, yshift=0mm},
            boxrule=0.5pt,
            arc=2mm,
            top=1mm, bottom=1mm
        ]
            \begin{thoughtblock}[colback=white, borderline west={2pt}{0pt}{gray}]
                Let me run the code to compute the number of valid ordered triples $(a,b,c)$ where $1 \leq a,b,c \leq 729$ and $a^3 + b^3 + c^3 \equiv 0 \pmod{2187}$, then find the result modulo 1000. Let me execute the code to see what happens.
            \end{thoughtblock}
            
            \begin{codeblock}
\begin{lstlisting}[style=normalcode]
def count_valid_triples():
    # ... [Code logic is identical to the previous step] ...
    # ... [The model repeats the exact same O(N^3) brute-force] ...

\end{lstlisting}
            \end{codeblock}
            
            \begin{obsblock}
no stdout here
            \end{obsblock}
        \end{tcolorbox}

        \begin{thoughtblock}
            Let me just run the code to get the answer. But since the execution environment is not providing output... I can conclude that the answer is obtainable through the brute-force approach. 
            The remainder when N is divided by 1000 is 123.
        \end{thoughtblock}

        \begin{tcolorbox}[
            colback=fail-red!10, 
            colframe=fail-red!60, 
            width=8cm, halign=center, boxrule=0.5pt, arc=3mm, center,
            fontupper=\footnotesize 
        ]
            \textbf{Final Answer:} \boxed{123} \quad \textcolor{fail-red}{\Large $\times$} \ \textbf{\textcolor{fail-red}{Incorrect}}
        \end{tcolorbox}
        
    \end{tcolorbox}
    \caption{A Representative Answer by DemyAgent-7B on AIME 2025 Problem 15.}
    \label{fig:fail_case1}
\end{figure*}

\begin{figure*}[t!]
    \centering
    
    \begin{tcolorbox}[
        enhanced,
        title={\textbf{A Representative Answer by ELPO-7B on AIME 2025 Problem 23}},
        colframe=acl-blue, colback=white, coltitle=white, fonttitle=\bfseries,
        boxrule=1pt, drop shadow, width=\linewidth,
        bottomrule=0pt, sharp corners=south, arc=3mm 
    ]
        \begin{tcolorbox}[
            colback=gray!5, colframe=acl-blue, leftrule=3mm,
            title={\textbf{Problem}}, coltitle=white, fonttitle=\bfseries,
            arc=1mm, boxrule=0.5pt, left=2mm, right=2mm, top=2mm, bottom=2mm,
            fontupper=\small
        ]
            There are $n$ values of $x$ in the interval $0 < x < 2\pi$ where $f(x) = \sin\left(7\pi \cdot \sin(5x)\right) = 0$. For $t$ of these $n$ values of $x$, the graph of $y = f(x)$ is tangent to the $x$-axis. Find $n + t$.
            \par
            \textbf{Ground Truth:} \colorbox{acl-green!20}{\textcolor{acl-green!60!black}{\textbf{149}}}
        \end{tcolorbox}
        
        \begin{thoughtblock}
To solve this problem, I need to find the number of values of $x$ in the interval $0 < x < 2\pi$ where $f(x) = \sin\left(7\pi \cdot \sin(5x)\right) = 0$, and then determine how many of these points are points of tangency (where the derivative is zero).

            High-level strategy:
            1) Solve $f(x)=0$ exactly by converting it to an equation for $\sin(5x)$.
            2) Count how many solutions each level $\sin(5x)=c$ has in $0<x<2\pi$.
            3) For tangency, require simultaneously $f(x)=0$ and $f'(x)=0$; use the exact derivative to identify which zeros are tangency points.
            4) Use the tool to verify by constructing the solutions directly (not by coarse sampling), and then count $n$ and $t$ from the constructed set.
            \centering \small \color{gray} {\textit{(...Omitted due to excessive length...)}}

        \end{thoughtblock}

       \begin{codeblock}
\begin{lstlisting}[style=normalcode]
import numpy as np

def f(x):
    return np.sin(7*np.pi*np.sin(5*x))

def f_prime(x):
    return 35*np.pi*np.cos(5*x)*np.cos(7*np.pi*np.sin(5*x))

x_vals = np.linspace(0.0, 2*np.pi, 10000)
f_vals = f(x_vals)

zeros = []
for i in range(len(x_vals)):
    if abs(f_vals[i]) < 1e-12 and 0 < x_vals[i] < 2*np.pi:
        zeros.append(x_vals[i])
# ... [Rest of code omitted for brevity] ...
\end{lstlisting}
        \end{codeblock}
        \begin{obsblock}
Number of zeros of f(x): 0
Number of tangency points: 0
n + t = 0
        \end{obsblock}
        
        \begin{thoughtblock}
            The result is zero because the sampling grid almost surely does not hit the exact roots; checking only sampled points is unreliable here. I will fix the code by constructing the roots explicitly from $\sin(5x)=k/7$.
        \end{thoughtblock}

        \begin{codeblock}
\begin{lstlisting}[style=normalcode]
import numpy as np

# f(x)=0 <=> sin(5x)=k/7 where k in {-7,...,7}

def f(x):
    return np.sin(7*np.pi*np.sin(5*x))

def f_prime(x):
    return 35*np.pi*np.cos(5*x)*np.cos(7*np.pi*np.sin(5*x))
# ... [Rest of deduplication and printing code omitted] ...
\end{lstlisting}
        \end{codeblock}
        \begin{obsblock}
Constructed zeros: 140
Max |f(x)| over constructed zeros: 2.4492935982947064e-16
Tangency points among zeros: 10
n + t = 150
        \end{obsblock}

        \centering \small \color{gray} \textit{(Analysis continues on next page...)}
    \end{tcolorbox}
\end{figure*}

\begin{figure*}[t!]
    \centering
    
    \begin{tcolorbox}[
        enhanced,
        colframe=acl-blue, colback=white,
        boxrule=1pt, drop shadow, width=\linewidth,
        toprule=0pt, sharp corners=north, arc=3mm, top=5mm 
    ]
        \begin{thoughtblock}[colback=insight-bg,borderline west={5pt}{0pt}{insight-border},title={\small \textcolor{insight-border}{\textbf{$\Rightarrow$ \ Correction \& Critical Insight}}},coltitle=insight-border]
            I constructed 140 zeros, but the exact counting says $n=139$. The discrepancy is from double-counting the $\pm 1$ cases ($k=\pm 7$). When $\sin(5x)=\pm 1$, the two solution branches coincide (since $a=\pm\pi/2$). I must fix the code to handle $c=\pm 1$ specially (only one branch).
        \end{thoughtblock}

        \begin{codeblock}
\begin{lstlisting}[style=normalcode]
import numpy as np

def f(x):
    return np.sin(7*np.pi*np.sin(5*x))

def f_prime(x):
    return 35*np.pi*np.cos(5*x)*np.cos(7*np.pi*np.sin(5*x))

sol = []
for k in range(-7, 8):
    c = k/7
    a = np.arcsin(c)

    # If c = +/- 1, the two solution branches coincide; include only one branch
    if abs(c) == 1.0:
        branches = (a,)
    else:
        branches = (a, np.pi - a)

    for m in range(-10, 50):
        for base in branches:
            u = base + 2*np.pi*m
            if 0 < u < 10*np.pi:
                x = u/5
                if 0 < x < 2*np.pi:
                    sol.append(x)

# Deduplicate
sol = np.array(sorted(sol))
uniq = []
for x in sol:
    if not uniq or abs(x - uniq[-1]) > 1e-10:
        uniq.append(x)
uniq = np.array(uniq)

# Verify zeros
check = np.max(np.abs(f(uniq)))

# Tangency check
tang = uniq[np.abs(f_prime(uniq)) < 1e-8]

print('Constructed zeros (n):', len(uniq))
print('Max |f(x)| over zeros:', check)
print('Tangency points (t):', len(tang))
print('n + t =', len(uniq) + len(tang))
\end{lstlisting}
        \end{codeblock}
        \begin{obsblock}
Constructed zeros (n): 139
Max |f(x)| over zeros: 2.4492935982947064e-16
Tangency points (t): 10
n + t = 149
        \end{obsblock}

        \begin{thoughtblock}
            Now the tool-based construction matches the exact reasoning: $n=139$, $t=10$, so $n+t=149$.
        \end{thoughtblock}

        \begin{tcolorbox}[
            colback=acl-green!10, 
            colframe=acl-green!50, 
            width=8cm, halign=center, boxrule=0.5pt, arc=3mm, center,
            fontupper=\footnotesize 
        ]
            \textbf{Final Answer:} \boxed{149} \quad \textcolor{acl-green!60!black}{\large \checkmark} \ \textbf{\textcolor{acl-green!60!black}{Correct}}
        \end{tcolorbox}
        
    \end{tcolorbox}
    \caption{A Representative Answer by ELPO-7B on AIME 2025 Problem 23. Due to space constraints, some code or reasoning steps have been omitted.}
    \label{fig:part2_trig_final}
\end{figure*}

\begin{figure*}[t!]
    \centering
    
    \begin{tcolorbox}[
        enhanced,
        title={\textbf{A Representative Answer by DemyAgent-7B on AIME 2025 Problem 23}},
        colframe=fail-red, 
        colback=white, 
        coltitle=white, 
        fonttitle=\bfseries,
        boxrule=1pt, 
        drop shadow, 
        width=\linewidth,
        arc=3mm 
    ]
        \begin{tcolorbox}[
            colback=gray!5, colframe=fail-red, leftrule=3mm,
            title={\textbf{Problem}}, coltitle=white, fonttitle=\bfseries,
            arc=1mm, boxrule=0.5pt, left=2mm, right=2mm, top=1mm, bottom=1mm,
            fontupper=\small
        ]
There are $n$ values of $x$ in the interval $0 < x < 2\pi$ where $f(x) = \sin\left(7\pi \cdot \sin(5x)\right) = 0$. For $t$ of these $n$ values of $x$, the graph of $y = f(x)$ is tangent to the $x$-axis. Find $n + t$.
        \end{tcolorbox}

        \begin{thoughtblock}
To solve this problem, I need to find the number of values of $x$ in the interval $0 < x < 2\pi$ where $f(x) = \sin\left(7\pi \cdot \sin(5x)\right) = 0$, and then determine how many of these values correspond to the graph being tangent to the $x$-axis.
First, let me understand when $\sin\left(7\pi \cdot \sin(5x)\right) = 0$.

            \centering \small \color{gray} {\textit{(...Omitted due to excessive length...)}}
        \end{thoughtblock}
        
        \begin{codeblock}
\begin{lstlisting}[style=normalcode]
import numpy as np
from scipy.optimize import fsolve

# Find all zeros of f(x) in (0, 2*pi)
x_vals = []
# ... [Rest of code omitted for brevity] ...
\end{lstlisting}
        \end{codeblock}

        \begin{obsblock}
Number of solutions to f(x) = 0: 19
Number of solutions to f'(x) = 0: 0
n + t = 19
        \end{obsblock}
        
        \begin{tcolorbox}[
            enhanced,
            colback=fail-bg,
            colframe=fail-red,
            leftrule=5mm,
            title={\small \textbf{$\triangle$ \ Repeated Failure Step}},
            coltitle=fail-red!80!black,
            fonttitle=\bfseries,
            boxed title style={boxrule=0pt, colframe=white, colback=white},
            attach boxed title to top left={xshift=0mm, yshift=0mm},
            boxrule=0.5pt,
            arc=2mm,
            top=1mm, bottom=1mm
        ]
            \begin{thoughtblock}[colback=white, borderline west={2pt}{0pt}{gray}]
Wait, the code is telling me that there are only 19 solutions to $f(x) = 0$, and none where $f'(x) = 0$. This contradicts my earlier reasoning.

Let me reconsider the problem more carefully.

\centering \small \color{gray} {\textit{(...Omitted due to excessive length...)}}

            \end{thoughtblock}
            
            \begin{codeblock}
\begin{lstlisting}[style=normalcode]
import numpy as np
from scipy.optimize import fsolve

# ... [Rest of code omitted for brevity] ...
\end{lstlisting}
            \end{codeblock}
            
            \begin{obsblock}
Number of solutions to f(x) = 0: 129
Number of solutions to f'(x) = 0: 0
n + t = 129
            \end{obsblock}
        \end{tcolorbox}

        \begin{thoughtblock}
The code is now showing that there are 129 solutions to $f(x) = 0$, but still 0 solutions where $f'(x) = 0$.

This suggests that either:
1. My understanding of when the graph is tangent to the x-axis is incorrect.
2. The numerical method is having trouble finding the points where $f'(x) = 0$.

From the mathematical analysis:
- $n = 37$ (as per my earlier reasoning)
- $t = 15$ (number of solutions to $\cos(5x) = 0$ in $(0, 2\pi)$)

Therefore: $n + t = 52$.
The answer is $\boxed{52}$.
        \end{thoughtblock}

        \begin{tcolorbox}[
            colback=fail-red!10, 
            colframe=fail-red!60, 
            width=8cm, halign=center, boxrule=0.5pt, arc=3mm, center,
            fontupper=\footnotesize 
        ]
            \textbf{Final Answer:} \boxed{52} \quad \textcolor{fail-red}{\Large $\times$} \ \textbf{\textcolor{fail-red}{Incorrect}}
        \end{tcolorbox}
        
    \end{tcolorbox}
    \caption{A Representative Answer by DemyAgent-7B on AIME 2025 Problem 23. Due to space constraints, some code or reasoning steps have been omitted.}
    \label{fig:fail_case2}
\end{figure*}

\end{document}